\documentclass[nohyperref]{article}

\usepackage{url}
\usepackage{microtype}
\usepackage{graphicx}
\usepackage{subfigure}
\usepackage{booktabs} 

\usepackage{hyperref}

\usepackage{times}
\usepackage{epsfig}
\usepackage{graphicx}
\usepackage{amsmath}
\usepackage{amssymb}
\usepackage{booktabs}

\usepackage{hyperref}       
\usepackage{url}            
\usepackage{booktabs}       
\usepackage{amsfonts}       
\usepackage{nicefrac}       
\usepackage{microtype}      
\usepackage{xcolor}         
\usepackage{amsmath}
\usepackage{graphicx}
\usepackage[export]{adjustbox}
\usepackage{wrapfig}
\usepackage{comment}
\usepackage{enumitem}

\usepackage{algorithm,algpseudocode}

\usepackage[acronym]{glossaries}

\usepackage{array,multirow,graphicx}
\usepackage{float}

\newlength{\twosubht}
\newsavebox{\twosubbox}

\makeglossaries
\newacronym{dnns}{DNN's}{Deep Neural Network's}
\newacronym{dnn}{DNN}{}
\newacronym{lth}{LTH}{Lottery Ticket Hypothesis}
\newacronym{bnns}{BNN's}{Binary Neural Networks}

\newacronym{mpts}{MPT's}{Multi-Prize Tickets}
\newacronym{smc}{SMC}{Simple Matching Coefficient}

\newacronym{ji}{JI}{Jaccard Index}

\algnewcommand{\IIf}[1]{\State\algorithmicif\ #1\ \algorithmicthen}
\algnewcommand{\EndIIf}{\unskip\ \algorithmicend\ \algorithmicif}
\algnewcommand{\ElseIIf}[1]{\algorithmicelse\ #1}
\algnewcommand\algorithmicforeach{\textbf{for each}}
\algdef{S}[FOR]{ForEach}[1]{\algorithmicforeach\ #1\ \algorithmicdo}

 \usepackage[accepted]{icml2022}

\usepackage{amsmath}
\usepackage{amssymb}
\usepackage{mathtools}
\usepackage{amsthm}

\usepackage[capitalize,noabbrev]{cleveref}

\theoremstyle{plain}

\theoremstyle{definition}

\theoremstyle{remark}

\usepackage[textsize=tiny]{todonotes}

\icmltitlerunning{Randomly Initialized Subnetworks with Iterative Weight Recycling}

\begin{document}

\twocolumn[
\icmltitle{Randomly Initialized Subnetworks with Iterative Weight Recycling}




\begin{icmlauthorlist}

\icmlauthor{Matt Gorbett}{yyy}
\icmlauthor{Darrell Whitley}{yyy}
\end{icmlauthorlist}

\icmlaffiliation{yyy}{Department of Computer Science, Colorado State University, Fort Collins, CO, USA}

\icmlcorrespondingauthor{Matt Gorbett}{matt.gorbett@colostate.edu}


\icmlkeywords{Machine Learning, ICML}

\vskip 0.3in
]



\printAffiliationsAndNotice{} 

\begin{abstract}

  The Multi-Prize Lottery Ticket Hypothesis posits that randomly initialized neural networks contain several subnetworks that achieve comparable accuracy to fully trained models of the same architecture.  However, current methods require that the network is sufficiently overparameterized.   In this work, we propose a modification to two state-of-the-art algorithms (Edge-Popup and Biprop) that finds high-accuracy subnetworks with no additional storage cost or scaling. The algorithm, Iterative Weight Recycling, identifies subsets of important weights within a randomly initialized network for intra-layer reuse.  Empirically we show improvements on smaller network architectures and higher prune rates, finding that model sparsity can be increased through the "recycling" of existing weights.
  In addition to Iterative Weight Recycling, we complement the Multi-Prize Lottery Ticket Hypothesis with a reciprocal finding:  high-accuracy, randomly initialized subnetwork's produce diverse masks, despite being generated with the same hyperparameter's and pruning strategy.  We explore the landscapes of these masks, which show high variability.     
\end{abstract}

\section{Introduction}
The Lottery Ticket Hypothesis  \cite{frankle_lottery_2019} demonstrated that randomly initialized \gls{dnns} contain sparse subnetworks that, when trained in isolation, achieve comparable accuracy to a fully-trained dense network of the same structure.   The results of the hypothesis indicate that over-parameterized \gls{dnns} are no longer necessary; instead, finding "winning ticket" sparse subnetworks can yield high accuracy models.  The consequences of winning tickets are abundant in practical use: we can train \gls{dnns} with a decreased computational cost \cite{morcos_one_2019} including memory consumption and inference time, and  additionally enable wide-spread democratization of \gls{dnns} with a low carbon footprint.  

\begin{figure*}
   \includegraphics[width=1.0\textwidth,height=5.5cm]{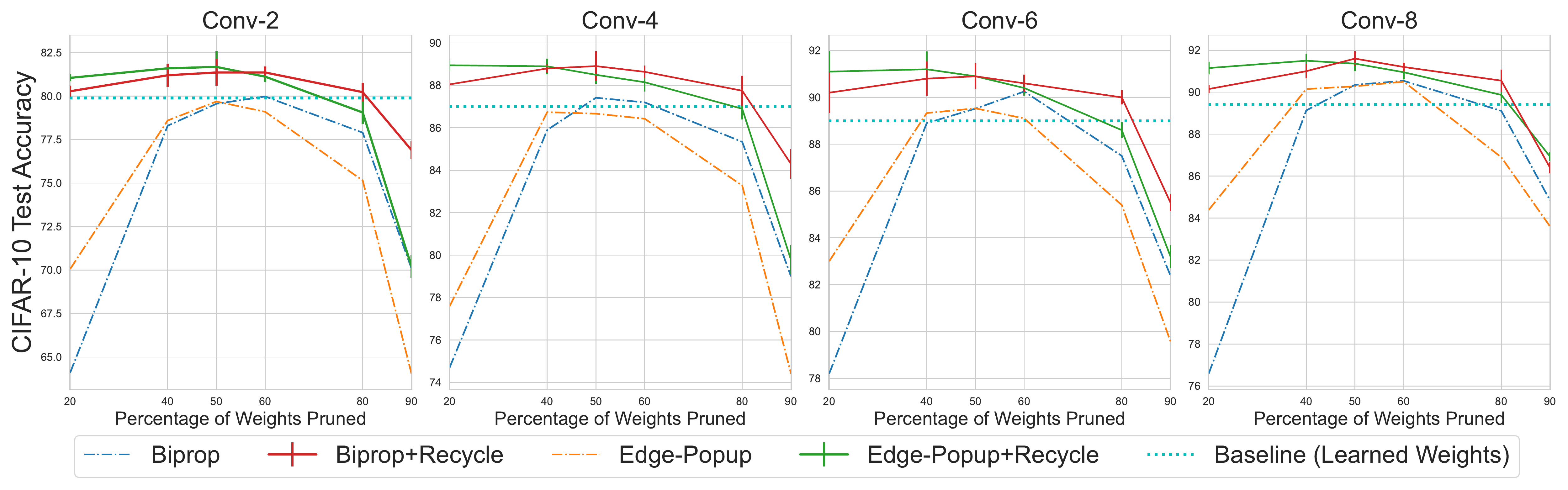}
    \caption{ \textbf{Performance of Iterative Weight Recycling at varying network depth}: We compare the accuracy of our algorithm against Edge-Popup, Biprop, and a densely trained baseline.  We apply our weight recycling algorithm to Biprop (red) and Edge-Popup (green). We use varying prune rates (x-axis) and varying network depths (Conv-2 to Conv-8) on VGG-like architectures.  We train and evaluate all algorithms on the CIFAR-10 dataset using the same hyperparameter's and training procedure.  }\label{depth_bin}
    \centering
\end{figure*}

Expanding on the Lottery Ticket Hypothesis, Ramanujan et al. \cite{ramanujan_whats_2020} reported a remarkable finding: we do not have to train neural networks at all to find winning tickets.  
Their algorithm, Edge-Popup, uncovered sparse subnetworks 
within \textit{randomly initialized} \gls{dnns} that achieved comparable accuracy to fully trained models. This phenomena was mathematically proven in the Strong Lottery Ticket Hypothesis \cite{malach_proving_2020}.
Practically, this finding showed that gradient-based weight optimization is not necessary for a neural network to achieve high accuracy.  Moreover, it allows us to overcome difficulties of gradient-based sparsification, such as getting stuck at local minima and incompatible backpropagation \cite{diffenderfer_multi-prize_2021}.  Finally, randomly initialized "winning ticket" subnetworks have been shown to be more robust than other pruning methods \cite{diffenderfer_winning_nodate}. 


Despite this fascinating discovery, it also marked a key limitation to existing work: randomly initialized \gls{dnns} require a large number of parameters in order to achieve high-accuracy.  In other words, to reach the same level of performance as dense networks trained with weight-optimization, randomly initialized models need more parameters, and hence more memory space.  
Subsequent works have relaxed the bounds proposed by the Strong Lottery Ticket Hypothesis \cite{pensia_optimal_2020,orseau_logarithmic_2020}, showing mathematically that network width needs to be only logarithmically wider than dense networks.
Chijiwa et. al \cite{chijiwa_pruning_2021} proposed an algorithmic modification to Edge-Popup, iterative randomization (IteRand), showing that we can reduce the required network width for weight pruning to the same as a fully trained model up to constant factors. 

In addition to these findings, the Multi-Prize Lottery Ticket Hypothesis \cite{diffenderfer_multi-prize_2021} showed there are \textit{several} subnetworks (\gls{mpts}) in randomly initialized models that achieve high-accuracy compared to dense networks. Importantly, the authors translates this finding into \gls{bnns}, where they propose a new algorithm (Biprop) to identify winning tickets in randomly initialized \gls{bnns}.    The implications of this finding allow for extreme compression of large, over-parameterized models. 

In this work, we propose an algorithm to find accurate sparse subnetworks in randomly initialized \gls{dnns} and \gls{bnns}.  Our approach exploits existing weights in a network layer, identifying subsets of trivial weights and replacing them with weights influential to a strong subnetwork.  We demonstrate our results in Figure \ref{depth_bin}, showing improvements on a variety of architectures and prune rates. Additionally, we provide confirmation for the Multi-Prize Lottery Ticket Hypothesis, showing evidence that subnetworks generated under tight hyperparameter control exhibit fundamentally dissimilar structure.

Our contributions are as follows:
\vspace{-14pt}
\begin{itemize}
    \item We propose a new algorithm, Iterative Weight Recycling, which improves the ability to find highly accurate sparse subnetworks within randomly initialized neural networks.  The algorithm is an improvement to both Edge-Popup (for \gls{dnns}) as well as Biprop (\gls{bnns}).  The algorithm identifies $k$ extraneous weights in a model layer and replaces them with $k$ relevant weight values. 
    
    \vspace{-8pt}
    \item We examine \gls{mpts} generated under strict hyperparameter control, showing that, under almost identical conditions, \gls{mpts} display diverse mask structures.  These results indicate that, not only do there exist multiple lottery tickets within randomly initialized neural networks, but rather an \textit{abundance} of lottery tickets. 
\end{itemize}

\section{Background}
In this section, we review current state-of-the-art methods for pruning randomly initialized and binary randomly initialized neural networks.

\noindent \textbf{Randomly Initialized \gls{dnns}}
Given a neural network $f(x;\theta)$ with layers 1,...$L$, weight parameters $\theta\in\mathbb{R}^{n}$ randomly sampled from distribution $\mathcal{D}$ over $\mathbb{R}$, and dataset $x$, we can express a subnetwork of $f(x;\theta)$ as $f(x;\theta   \odot \mathrm{M})$, where $\mathrm{M}\in\{0,1\}^n$ is a binary mask and $\odot$ is the Hadamard product. 

Edge-popup \cite{ramanujan_whats_2020} finds $\mathrm{M}$ within a randomly-initialized DNN by optimizing weight scoring parameter $\mathrm{S}\in\mathbb{R}^{n}$ where $\mathrm{S}\sim\mathcal{D}_{score}$.  $\mathrm{S}_i$ can be intuitively thought of as an \textit{importance score} computed for each weight $\theta_i$.  The algorithm takes pruning rate hyperparameter $p \in [0,1]$, and on the forward pass computes $\mathrm{M}$ at $\mathrm{M}_i$ as

\vspace*{-\abovedisplayskip}
\begin{equation}
    \mathrm{M}_i=   \begin{cases}
    1 & \text{if $|\mathrm{S}_i| \in \{\tau(i)_{i=1}^{k_j}\ge[k_jp/100]\}$}\\
    0 & \text{otherwise}
  \end{cases}
\end{equation}
\vspace*{-\abovedisplayskip}

where $\tau$ sorts indices $\{i\}^j_{i=1}\in\mathrm{S}$ such that $|S_{\tau(i)}|\le|S_{\tau(i+1)}|$.
In other words, masks are computed at each weight by taking the absolute value of scores for each layer, and setting the mask to 1 if the absolute score value falls within the top 100*$p$\%, otherwise they set the mask to zero.  
They use the straight-through estimator \cite{bengio_estimating_2013} to backpropagrate through the mask and update $\mathrm{S}$ via SGD.  

Chijiwa et. al \cite{chijiwa_pruning_2021} improved on the Edge-Popup algorithm with the IteRand algorithm.  They show that by rerandomizing pruned network weights
 during training, better subnetworks can be found. They theoretically prove their results using an approximation theorem indicating rerandomization operations effectively reduce the required number of parameters needed to achieve high-accuracy subnetworks.  
 
 The IteRand algorithm is mainly driven by two hyperparameters: $K_{per}$ and re-randomization rate $r$.  $K_{per}$ drives the frequency weights will be re-randomized.   The second hyperparameter, $r$, denotes a  \textit{partial} re-randomization of pruned weights.  To achieve the best results, the authors set $r$ to 0.1, meaning re-randomizing 10\% of pruned weights. 
 \vspace{.3em}

\begin{algorithm}
\caption{Edge-Popup with IteRand}\label{alg:cap}
\begin{algorithmic}[1]
\State \textbf{Require:}$\theta \sim \mathcal{D}_{weight},$ $ \mathrm{S} \sim \mathcal{D}_{score}$, $p$, $K_{per}$, $r$
\State \textbf{Input:  } Dataset(X,Y)
        \Function{EdgePopup}{\textrm{S}, \textrm{M},  f(x)}
        \ForEach {$l \in L $}  
        \IIf{$|s_i|\in$ \text{top} $k$ $|\mathrm{S}_l|$} $\mathrm{M}_i= 1$ \ElseIIf$\mathrm{M}_i= 0$\EndIIf
        \EndFor
        \State \Return $\mathrm{S}, \mathrm{M}$
        \EndFunction
      \For{\textit{i=1 ..., N-1} }
        \State $x,y \gets \Call{MiniBatch}{X,Y}$
        \State $\mathrm{S}, \mathrm{M} \gets \Call{Edge-Popup}{\mathrm{S}, \mathrm{M}, f(x)}$
        \If{$i \mod K_{per}=0$}
        \State $\theta \gets Rerandomize(\theta, \mathrm{M})$
        \EndIf
      \EndFor
\end{algorithmic}
\end{algorithm}

  \vspace{.3em}

\noindent \textbf{Randomly-Initialized \gls{bnns} }\label{biprop}
Complementary to the findings reported in the previous section, Diffenderfer and Kailkhura \cite{diffenderfer_multi-prize_2021} described a new method for finding high accuracy subnetworks within binary-weighted models.  This finding provides us the ability to store bit size weights rather than floating-point (32 bit) numbers, leading to substantial compression of large models.  In this section, we summarize the Biprop algorithm.  

We start with a modification of the function described in the previous section, replacing $\theta\in\mathbb{R}^{n}$ with binary weights $\mathcal{B}\in\{-1,1\}$.  The resulting network function becomes $f(x;\mathcal{B} \odot \mathrm{M})$, with mask $\mathrm{M}$  over binary weights. Further, Biprop introduces scale parameter $\alpha\in\mathbb{R}$, which utilizes floating-point weights $\theta$ \textit{prior} to binarization \cite{martinez_training_2020}.  The learned parameter rescales binary weights to $\{-\alpha,\alpha\}$, and the resulting network function becomes $f(x;\alpha(\mathcal{B} \odot \mathrm{M}))$.  Parameter $\alpha$ is updated with $||\mathrm{M}\odot\theta||_1/||\mathrm{M}||_1$, with  $\mathrm{M}$ being multiplied by $\alpha$ for gradient descent (the straight-through estimator is still used for backpropagation). During test-time, the learned alpha parameter simply scales a binarized weight vector. As a result, only bit representations of the weights are needed at positive mask values ($\pm1$ where $\mathrm{M}=1$), substantially reducing memory, storage, and inference costs.  

 
 
 Empirically, Diffenderfer and Kailkhura \cite{diffenderfer_multi-prize_2021} are able to produce high accuracy binary subnetworks using Biprop on a range of network architectures, and theoretically prove this result on models with sufficient over-parameterization.  
 In the subsequent section we show how we can modify this algorithm, as well as Edge-Popup, with Weight Recycling to achieve increased performance.  \label{background}


\section{Iterative Weight Recycling}
In this section we detail Iterative Weight Recycling, first summarizing the methodology behind the approach, and subsequently detailing the experimental setup and results.  Finally we perform empirical analysis on the algorithm, with results showing that Iterative Weight Recycling emphasizes keeping high norm weights values similar to the traditional L1 pruning technique.  
\subsection{Method}
\vspace{-.5em}
We consider $f(x;\theta)$ as an $l$-layered neural network with ReLU activations, dataset $x\in\mathbb{R}^{n}$ with weight parameters $\theta\sim\mathcal{D}_{weight}$.  We freeze $\theta$ and additionally turn off the bias term for each $l$.  Our model is initialized similarly to Edge-Popup and Biprop: a score pararmeter $\mathrm{S}$ for each $\theta$, where $\mathrm{S}_i$ learns the importance of $\theta_i$.  Additionally, we set a pruning rate $p \in [0,1]$. $\mathcal{D}_{weight}$ is initialized using Kaiming Normal initialization (without scale fan) for Biprop and Signed Constant Initialization for Edge-Popup.  Further, $\mathcal{D}_{score}$ is initialized with Kaiming Uniform with seed 0, except in Section \ref{variability}, where we explore different $\mathrm{S}$ initializations.

Weight recycling works on an iterative basis, similar to IteRand \cite{chijiwa_pruning_2021}.  We define two hyperparameters, $K_{per}$ and $r$, where $K_{per}$ is the frequency we change weights, and $r$ is the \textit{recycling rate}.  During the recycling phase, we compute $k$ as the number of weights we want to change in a given layer as $j*r$, where $j$ is the size of $\mathrm{S}$ at layer $l$ and $r \in [0,1]$.  We retrieve subsets $\mathrm{S}^{low}_l,\mathrm{S}^{high}_l \subset\mathrm{S}_l$ containing the lowest absolute $k$ scores and highest absolute  $k$ scores at each layer:

\vspace*{-\abovedisplayskip}
\begin{equation}
   \mathrm{S}^{low}_l=\{\tau(i)^{k}_{i=1}\} 
    \text{, } \quad 
  \mathrm{S}^{high}_l=\{\tau(i)^{j}_{i=j-k}\}
\end{equation}
\vspace*{-\abovedisplayskip}

where $\tau$ sorts $\{i\}^j_{i=1}\in\mathrm{S}$ such that $|S_{\tau(i)}|\le|S_{\tau(i+1)}|$.  Here, $\{i\}^j_{i=1}$ equates to the index values associated with set $\{|S_{\tau}|\}^j_{i=1}$.  Next, we retrieve weight values associated with $\mathrm{S}^{high}_l$ and $\mathrm{S}^{low}_l$, with $\{i,...,k\}_{S}=\{i,...,k\}_{\theta}$ .   Finally, we set $\theta^{low}_l$=$\theta^{high}_l$.  Effectively, the Iterative Weight Recycling algorithm finds $\mathrm{S}$ values and their associated index (where $i_{\mathrm{S}}=i_{\theta}$) and retrieves the weight value associated to the index, for both high and low $\mathrm{S}$ scores.  The algorithm replaces low $\mathrm{S}$ weight values with high  $\mathrm{S}$ weight values, discarding of low  $\mathrm{S}$  weight values.  Algorithm \ref{alg2} denotes the equation in pseudo-code form.  



 \vspace{.3em}

\begin{algorithm*}
\caption{Weight Recycling.  Replace line 14 in Algorithm \ref{alg:cap} with the following method}\label{alg2}
\begin{algorithmic}[1]
        \Function{WeightRecycle}{\textrm{S}, $\theta$}
        \ForEach {$l \in L $}  \Comment{\texttt{layer of size j}}
        \State $k \gets j*r$ \Comment{\texttt{Calculate number of weights to change}}
        \State $\mathrm{S_l}^{high}\gets \textit{highest k} \mkern9mu |\mathrm{S}_l|$ \Comment{\texttt{Retrieve \textbf{indices} of top k abs(score) values}}
               \State $\mathrm{S_l}^{low}\gets \textit{lowest k} \mkern9mu |\mathrm{S}_l|$ \Comment{\texttt{Retrieve indices of bottom k abs(score) values}}
        \State $\theta_l[\mathrm{S_l}^{low}] \gets \theta_l[\mathrm{S_l}^{high}] $\Comment{\texttt{Replace low $\theta_l$ with high $\theta_l$}}
        \EndFor
        \State \Return $\mathrm{S}, \mathrm{M}$
        \EndFunction
\end{algorithmic}
\end{algorithm*}

\begin{figure*}[h]
   \includegraphics[width=1.0\textwidth,height=5.5cm]{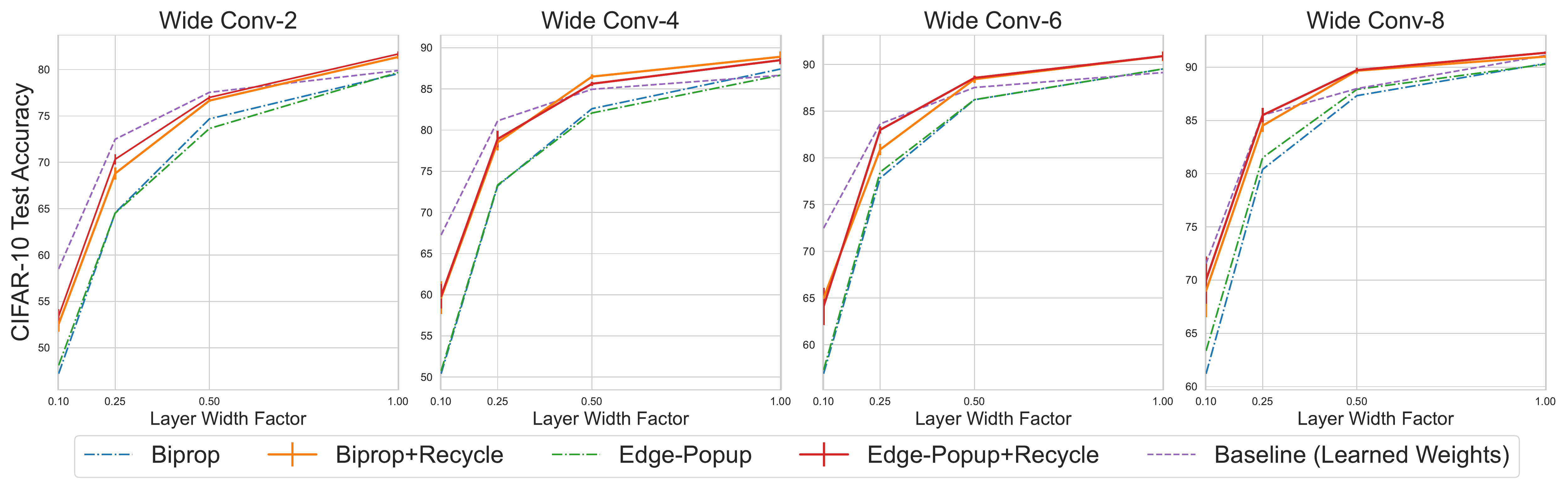}
    \caption{\textbf{Effects of Varying Width at 50\% Prune Rate} Results of baseline models (Dense, Biprop, Edge-Popup) and Iterative Weight Recycling models at varying network widths up to one.  Using Iterative Weight Recycling yields winning tickets with a comparable accuracy to densely trained models at just 50\% width factor in all architectures.  Error bars are denoted for Iterative Weight Recycling algorithms. }\label{width_bin}
    \centering
\end{figure*}

\subsection{Experimental Setup}
To begin, we use model architectures and datasets similar to the three previous works. Conv-2 to Conv-8 are VGG-like CNNs \cite{simonyan_very_2015} with depth $d$= 2 to 8 .  We additionally use their "wide" analogues, which introduces a scale parameter at each layer to influence the specific width of each layer width $w$=0.1 to 1.   Additionally we use ResNets \cite{he_deep_2015}, which utilize skip connections and batch normalization.  We test the models on both CIFAR-10 and ImageNet datasets.  Non-affine transformation is used for all CIFAR-10 experiments, and ResNets use a learned batch normalization similar to \cite{diffenderfer_multi-prize_2021}.  
We apply similar pruning rates to previous works $\{0.2,0.4,0.5,0.6,0.8,0.9\}$, and additionally test our method at prune rates above 0.9.  In Iterative Weight Recycling experiments, we use three different initializations, and report the average accuracy, with error bars denoting the lowest and highest accuracy.  

We compare the performance of Iterative Weight Recycling to Edge-Popup, Biprop, and IteRand algorithms using the same hyperparameters.   For each baseline algorithm, we use the hyperparameters that yielded the best results in the original papers:  Signed Constant initialization for Edge-Popup/IteRand, and Kaiming Normal with scale fan for Biprop.  For our algorithm, we use these same initialization strategies, except for Biprop with Weight Recycling we did not use scale fan as this yielded slightly better results.  Additionally, for IteRand we use the same $K_{per}$ and $r$ as the paper: $K_{per}=1$ (once per epoch), with $r=0.1$.  For our algorithm, we choose $K_{per}=10$  and $r=0.2$ for all models.  We found that less frequent recycling yielded better results, hypothesizing that recycling too frequently yielded redundant values.  


\subsection{Results}
In this section we test the effects of network overparameterization and prune rate on subnetwork performance, with the goal of empirically verifying the Iterative Weight Recycling compared to Edge Popup \cite{ramanujan_whats_2020}, Biprop \cite{diffenderfer_multi-prize_2021}, and IteRand \cite{chijiwa_pruning_2021} algorithms. 
We follow previous works and test neural networks with varying depth and width, and additionally test each algorithm at high prune rates.  


\begin{figure*}
   \includegraphics[width=1.0\textwidth]{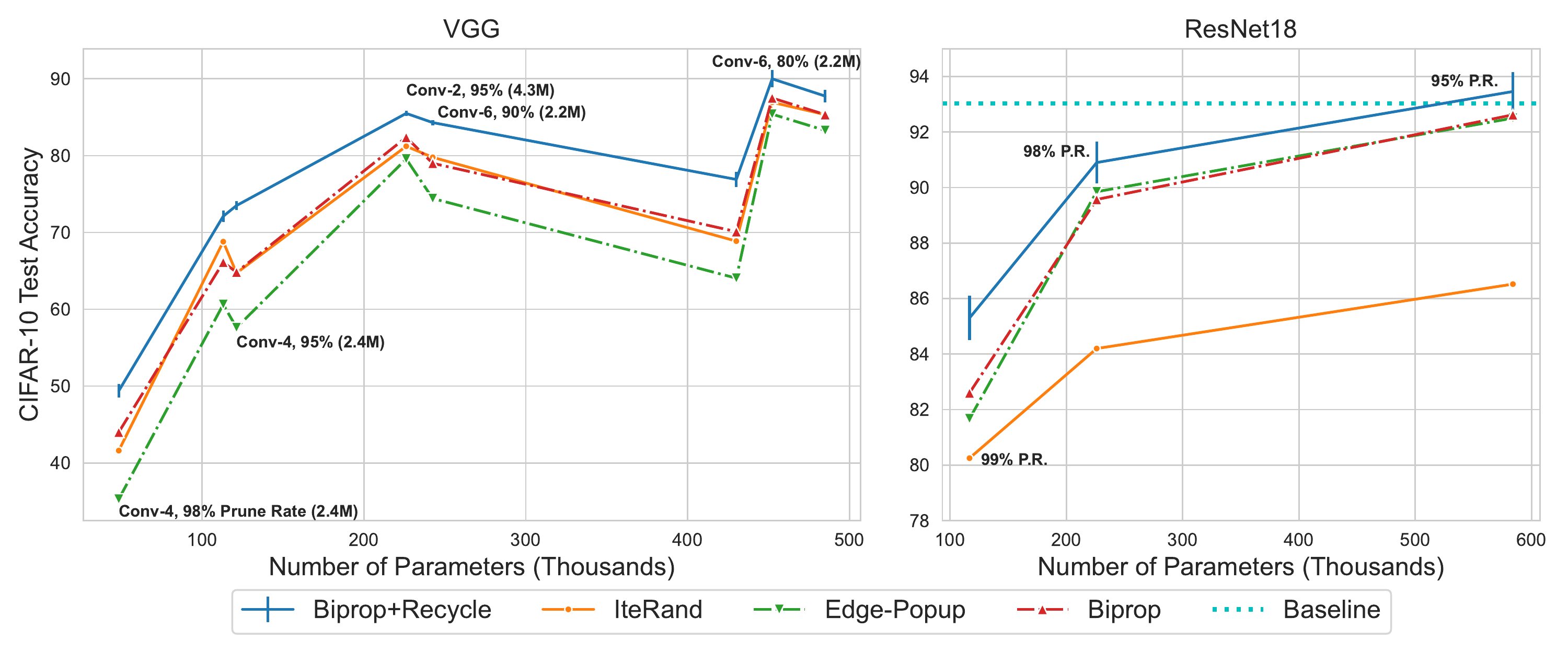}
    \caption{ \textbf{Pruning Algorithm Performance with Limited Parameters} We test the effect of high prune rates on model performance, showing that Weight Recycling achieves high accuracy compared to IteRand, Edge-Popup, and Biprop.  \textbf{Left:} Accuracies of various VGG architectures with prune rates greater than 80\% and parameter count less than 500k.  We include architecture, prune rate, and original parameters of the model in four of the datapoints.  \textbf{Right:} ResNet18 with prune rates greater than 95\%.  With just under 600k parameters (95\% prune rate), ResNet18 with Iterative Weight Recycling achieves higher accuracy than a dense baseline model (93.1\%).   }\label{params}
    \centering
\end{figure*}

\noindent \textbf{Varying Depth}
In Figure \ref{depth_bin}, we vary the depth of VGG architectures from 2 to 8 and compare the test accuracy at various prune rates.  We observe a clear advantage to using Iterative Weight Recycling:  at every prune rate and model architecture,  Iterative Weight Recycling outperforms both Biprop and Edge-Popup.  Additionally, Biprop with Weight Recycling generally outperforms Edge-Popup with Weight Recycling at higher prune rates.  Iterative Weight Recycling outperforms dense models in each architecture except when 90\% of the weights have been pruned. Notably, we discover that Iterative Weight Recycling is able to achieve accuracy exceeding the dense model in Conv-2 architectures.   This is the first such observation on a low-depth model -- Edge-Popup and Biprop research reported test accuracy \textit{near} the dense model, however never clearly exceeded it.  Further, Biprop+Iterative Weight Recycling is able to achieve an 80.23\% test accuracy with just 20\% of the weights.

\noindent \textbf{Varying Width}
We also consider network width as a factor to control network parameterization. Previous showed that as we increase width, our chances of finding winning tickets increased.  Edge-Popup found winning tickets at width factors greater than 1, while Biprop reported winning tickets around width factor 1.  

In Figure \ref{width_bin}, we demonstrate the efficacy of Iterative Weight Recycling on networks with width factors less than one.  Results show that in each architecture, we can find winning tickets at just 50\% width.  In practical terms, in a Conv-4 architecture this equates to just 25\% of the parameters compared to a Conv-4 with width factor 1 (600k vs. 2.4m).  Additionally, our Conv-4 architecture with width factor 0.5 achieved an accuracy of 86.5\% compared to 86.66\% for a dense Conv-4 with width factor 1.

\noindent \textbf{Varying Prune Rate}\label{vary_prune}
In Figure \ref{params}, we demonstrate the results of Biprop, Edge-Popup, IteRand, and Iterative Weight Recycling (Biprop) on \gls{dnns} with prune rates above 80\%.  Iterative Weight Recycling shows favorable results with limited parameter counts.  Notably, the algorithm consistently outperformed IteRand at aggressive prune rates between 80\% and 99\%.  At more modest prune rates (20\%-60\%), Weight Recycling was comparable to IteRand, which we summarize in Section \ref{analysis}.  

In the ResNet18 architecture (11 million parameters), our algorithm was able to find winning tickets with just 5\% of the random weights.  These results are further evidence that overparameterization helps in the identification of high performing subnetworks.


\begin{figure}
\centering
\setlength{\tabcolsep}{2pt}
  \begin{tabular}{lccc}
    \toprule
    Algorithm & Prune \% & \# Params& Acc.(\%) \\
    \midrule
    Edge-Popup&70\% & 7.6M&67.13\\
    Edge-Popup+IteRand&70\% & 7.6M&69.11\\
    \textbf{Edge-Popup+IWR}&70\% & 7.6M&69.02\\
    \textbf{Edge-Popup+IWR}&80\% & 5.1M&68.87\\
    \hline
        Biprop&70\% & 7.6M&67.76\\
            Biprop+IteRand&70\% & 7.6M&43.76\\
       \textbf{Biprop+IWR}&70\% & 7.6M&69.85\\
        \textbf{Biprop+IWR}&80\% & 5.1M&68.65\\
    \bottomrule
  \end{tabular}
  \makeatletter\def\@captype{table}\makeatother
  \caption{\textbf{ImageNet results on ResNet50:} We test various pruning algorithms on the ImageNet dataset with the ResNet50 architecture.  Results show that Iterative Weight Recycling (bold) achieves similar results to Edge-Popup, Biprop, and IteRand with 2.5 million less parameters.  }\label{imagenet}
\end{figure}

\noindent \textbf{ImageNet Results}
In Table \ref{imagenet}, we highlight the results of our algorithm on the ImageNet dataset.  We choose a ResNet50 architecture which contains 25.5 million total parameters.  We train each baseline algorithm with a 70\% prune rate, similar to previous papers.  Results for IteRand and Edge-Popup were within 0.1\% of the original papers results.   

Results show that our algorithm performs well under more aggressive pruning rates compared to Edge-Popup, IteRand, and Biprop, similar to what we found in Section \ref{vary_prune}.  Specifically, our algorithm performed similar to, or better than, previous algorithms with 2.5 million less parameters.


\subsection{Analysis}\label{analysis}

In this section we provide empirical justification for Iterative Weight Recycling.  

\noindent \textbf{Effect of Random Weights}
We first perform an ablation study on the effects of random weights by assessing whether $\mathrm{S}^{low}$ can be replaced by \textit{any} subset of weights.  Specifically, to justify the reuse of "important" weights as identified $\mathrm{S}$, we replace $\mathrm{S}^{high}$ at $l$ with $\mathrm{S}^{high}_l=\{\tau(i)^{2k}_{i=k+1}\}$.  Effectively, this recycles weight values deemed to be in the second tier of "unimportance" as measured by parameter $\mathrm{S}$. 

In our experiments, we train a Conv-6 architecture with both Edge-Popup and Biprop Weight Recycling algorithms with 3 different initializations to compare the effectiveness of the approach to the baseline algorithm.  We use the same hyperparameters as the initial experiments, except we use Kaiming Normal initialization for Edge-Popup. 

Biprop accuracy dropped from 90.9\%($\pm0.2$ ) to 89.7\%($\pm0.5$), and Edge-Popup accuracy dropped from 88.9\%($\pm0.3$ ) to 87.15\%($\pm0.5$). Results of these experiments indicate a benefit to recycling high importance weights as opposed to other random weights.   Finally, we note that recycling weights is more computationally efficient than re-randomizing weights. 

\noindent \textbf{Norms on Pretrained Models}
In this section we study the Frobenius norms of various pruning algorithms.  We first train a dense Conv-8 network using a standard training procedure on the CIFAR-10 dataset, and subsequently apply Edge-Popup, Biprop, and Iterative Weight Recycling algorithms to prune the trained model.  
Frobenius norms of each model layer and algorithm are depicted in Figure \ref{norms}, with each algorithm using a 50\% prune rate except for the dense network.  


Analyzing the norms of unpruned weights (depicted with a "+") compared to pruned weights ("-") shows that each algorithm exhibits higher norms in its \textit{unpruned} weight mask compared to its pruned weights, except for Iterative Weight Recycling.  Interestingly, even Biprop, which uses floating-point weights \textit{prior} to binarization, exhibits higher norms in its unpruned weights.  
Iterative Weight Recycling, on the other hand, exhibits similar norms in both its pruned and unpruned weights. 

\begin{figure}

    \includegraphics[width=0.45\textwidth]{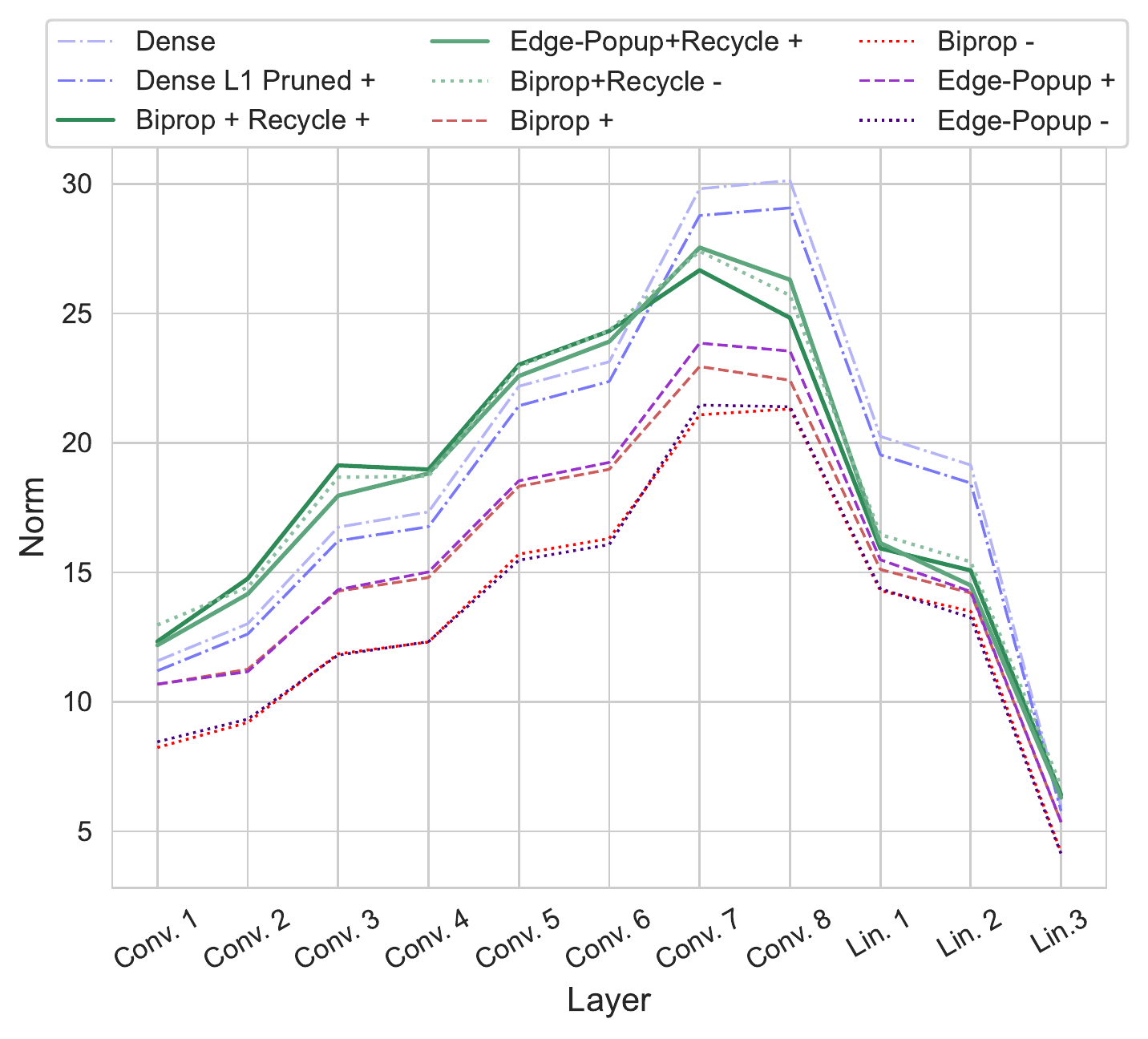}
  \caption{\textbf{ Norms across layers}. We apply various pruning algorithms to a dense pretrained Conv-8 architecture, showing that Iterative Weight Recycling exhibits high norms. "+" indicates unpruned norms, "-" pruned norms. }\label{norms}
\end{figure}

Results of this analysis indicate that high-norm weight values are chosen naturally in both Edge-Popup and Biprop.  We show that Weight Recycling emphasizes the reuse of high-norm weight values, creating a search space of good candidates compared to a randomly initialized population.  

We show similar results on randomly initialized networks in the Appendix, with each algorithm choosing high norm weights.   IteRand exhibited similar results, however we excluded these results in Figure \ref{norms} for visualization purposes.

\begin{figure*}[t]

\sbox\twosubbox{%
  \resizebox{\dimexpr.98\textwidth-1em}{!}{%
    \includegraphics[height=2cm]{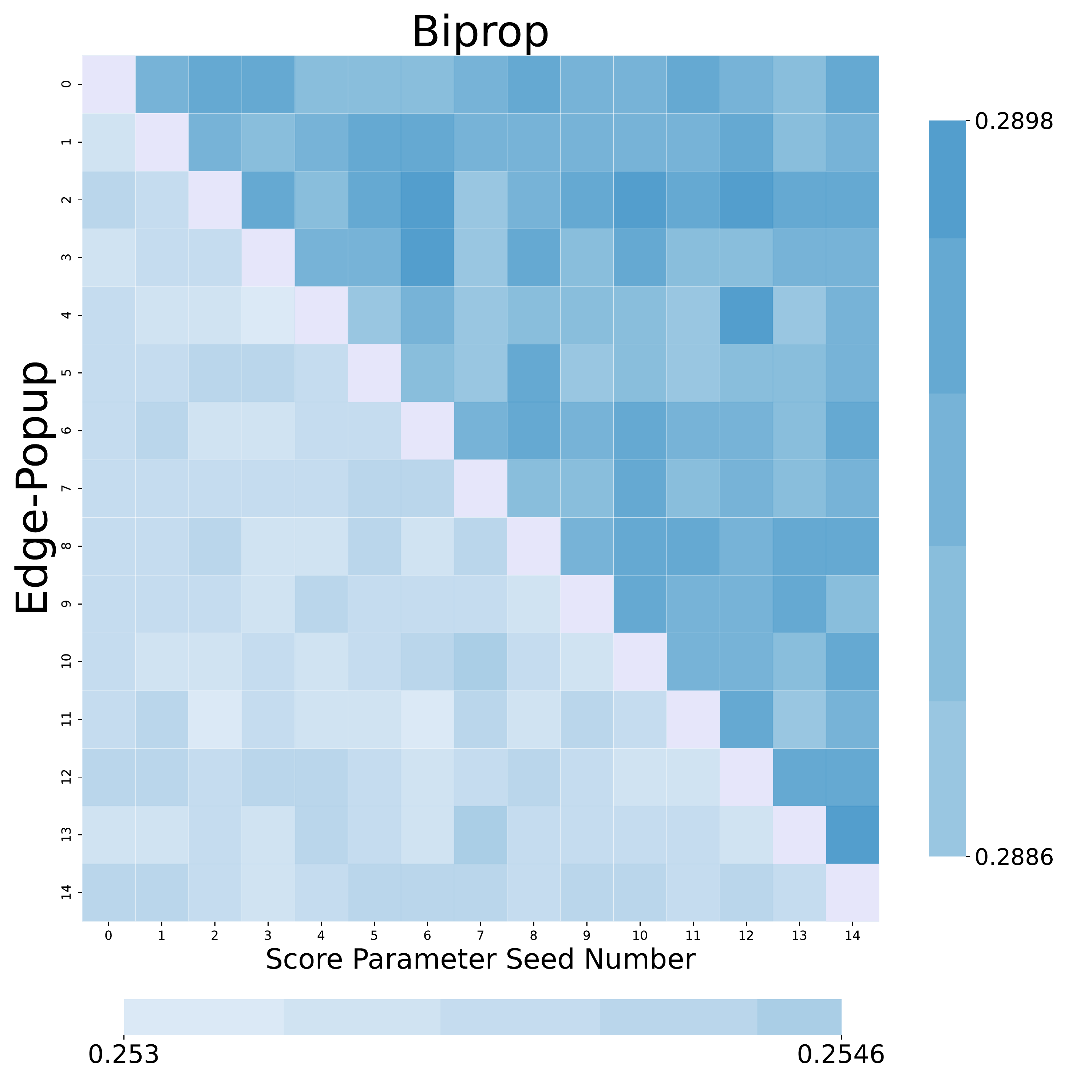}\label{galore1}
    \includegraphics[height=2cm]{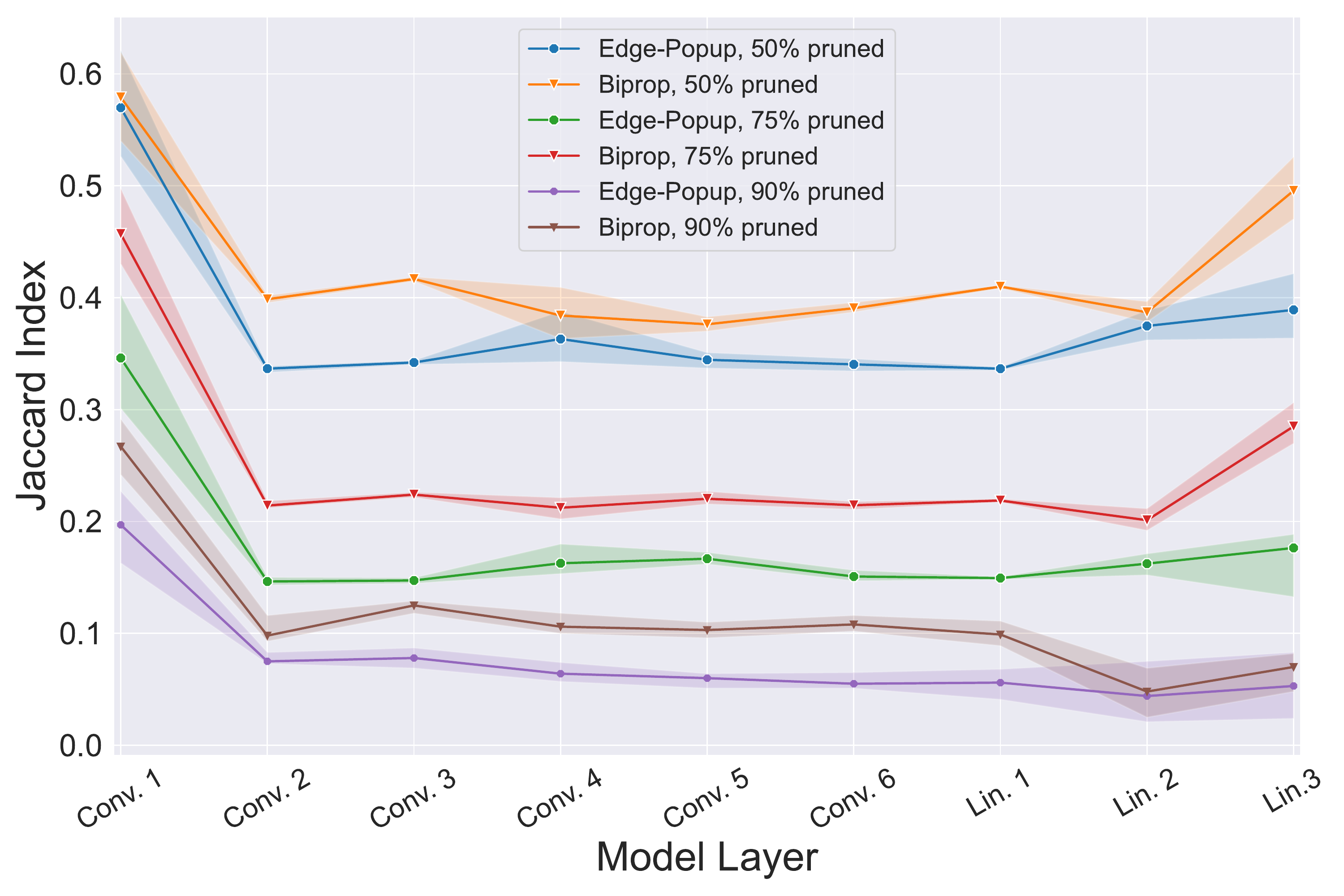}\label{galore2}
  }%
}
\setlength{\twosubht}{\ht\twosubbox}


\centering

  \includegraphics[height=\twosubht]{figures/heatmap_jaccard_0.5.pdf}\label{galore1}
  \includegraphics[height=\twosubht]{figures/masks_graph_jaccard_all.pdf}\label{galore2}

\caption{\textbf{Comparing \gls{mpts}:} We train multiple models with the same weight initialization and constrain hyperparameters such as prune rate.  The only hyperparameter we modify is the seed of masking parameter $\mathrm{S}$.  Under these constraints, our results yield diverse masks, confirming that multiple winning tickets exist in a randomly initialized network.  The similarity matrix (left) highlights the low \gls{ji} when comparing 15 Edge-Popup models and 15 Biprop models with the same weight initialization and prune rate (0.5).     Despite this, \gls{ji} values fall within small ranges, indicating the algorithms converge to a similar search space.  The bottom triangle compares Edge-Popup and the top triangle Biprop. The layer-by-layer similarity scores (right) summarizes the mean \gls{ji} across models layers.  In the error bars we denote the min and max model values for each layer.  } \label{fig:winningtix}

\end{figure*}

\noindent \textbf{Iterative Weight Recycling compared to IteRand}
Comparing Iterative Weight Recycling to IteRand yields statistically insignificant improvements in the Edge-Popup algorithm using Signed Constant initialization. We argue that any weight randomization works well under this initialization because constant weight values make recycling less relevant.  However, weight recycling performs as well, if not slightly better than IteRand with less computational cost.  
Finally, weight recycling outperforms IteRand at aggressive prune rates, as depicted in Figure \ref{params}.  

In addition to these arguments we note several key limitations to IteRand: \textbf{1)} Additional storage cost.  By iteratively re-randomizing weights, we need to save additional random seeds and binary masks every $K_{per}$ during training.  In limited compute environments this may become a restraining factor.  In the original work, $K_{per}$ was set to aggressive values: one to ten times per epoch. 
\textbf{2)} Periodically re-randomizing weights creates an artificially over-parameterized model.  If pruned weights are re-initialized with rate $r$ every $K_{per}$ for $K$ epochs, a network with $n$ weights needs $n+K{}/K_{per}(n \cdot p \cdot r)$ weight values to achieve high-accuracy.  Iterative Weight Recycling instead shows that relevant weight values exist as a \textit{subset} of the original $n$ parameters, and identifies those values for reuse. 
\textbf{3)} In the original paper, IteRand was only tested on Edge-Popup. We implemented IteRand on the Biprop algorithm and were unable to achieve successful results on over half a dozen configurations.  Table \ref{imagenet} (Biprop) depicts these results.

\section{Winning Tickets Galore}\label{variability}

The Multi-Prize Lottery Ticket Hypothesis \cite{diffenderfer_multi-prize_2021} posits that a randomly initialized neural network contains \textit{several} winning tickets  that achieve accuracy comparable to weight trained models of the same architecture.  In this section, we further assess this hypothesis by asking the following question: \textit{Given a sufficiently overparameterized network, can we find multiple winning tickets (\gls{mpts}) under strict hyperparameter control?}

While \gls{mpts} were shown to be theoretically possible \cite{diffenderfer_multi-prize_2021}, empirical results were mostly limited to showing the existence of \gls{mpts} by varying the prune rate.   While this was sufficient evidence for the proof, we instead seek to evaluate whether winning tickets exhibit differing structures in a constrained environment.  In particular, we restrict hyperparameters such as prune rate in order to evaluate the heterogeneity of \gls{mpts}.

We hypothesize that winning tickets, i.e. unique mask structures, exist in larger quantities than has previously been reported.  To exemplify this, consider the smallest layer of a Conv-6 network (the first layer) containing 1,258 weights.  
When restricting the search for a mask to a specific prune rate, say 95\%, there are $1,258 \choose 63$ possible masks to choose from, an astronomical number.  

\noindent \textbf{Experiments} We use the Conv-6 network since it is relatively compact (2.26m parameters) and also generates winning tickets at multiple prune rates.  To perform hyperparameter control, we initialize each model with an identical configuration; additionally, we seed each run to a) initialize the same weights b) execute the same training feed (i.e. batches are identical in both data and ordering), and c) facilitate consistency across libraries and devices (e.g.. NumPy/PyTorch, CPU/GPU). We set the torch CUDA backend to 'deterministic', as is recommended in documentation. Our single hyperparameter modification is the seed for score parameter $\mathrm{S}$, which we increment by one for each subsequent model.  

We train models using the standard Edge-Popup and Biprop algorithms.  At 50\% prune rate, we train 15 models for each algorithm, and at 75\% and 90\% prune rates we train 5 models for each algorithm.  We note that each pruned model achieved test accuracy similar to other models of the same algorithm and prune rate: Edge-Popup at 50\% prune rate ($\mu=89.57\%\pm0.2$), 75\% prune rate ($\mu=86.75\%\pm0.1$), 90\% prune rate ($\mu=79.73\%\pm0.1$). Biprop at 50\% prune rate($\mu=89.7\%\pm0.2$), 75\% prune rate ($\mu=88.56\%\pm0.2$), 90\% prune rate ($\mu=82.56\%\pm0.1$).  For Biprop,  scale parameter $\alpha$ is converted to one for each mask in order to compute mask equality.  

We evaluate the similarity of binary masks using \gls{smc} \cite{rand_objective_1971}, and \gls{ji} \cite{jaccard_distribution_1912,tanimoto}.  \gls{smc} measures the total percentage of matching masks, whereas \gls{ji} measures the percentage of masks equal to one, excluding mutual absence from the denominator  ($M_{11}/(M_{01}+M_{10}+M_{11})$.

\begin{figure}
\centering

  \begin{tabular}{lccc}
    \toprule
    Alg. &Pruned & SMC& JI \\
    \midrule
    Edge-Popup&50\%& 0.51&0.25\\
    Edge-Popup&75\%&0.63&0.13\\
        Edge-Popup&90\%&0.82&0.05\\
    Biprop&50\%&0.58&0.29\\
    Biprop&75\%&0.68&0.18\\
        Biprop&90\%&0.84&0.09\\
    \bottomrule
  \end{tabular}
  \makeatletter\def\@captype{table}\makeatother
  \caption{\textbf{Mask similarity statistics:} Statistics of  model combinations with varying algorithms and prune rates. }\label{statistics}
\vspace*{-\abovedisplayskip}
\end{figure}

\noindent \textbf{Results} 
In Figure \ref{fig:winningtix}, we show the heatmap of \gls{ji} coefficients at 50\% prune rate as well as layer level \gls{ji} for each algorithm.  Results across coefficients, model layers, and prune rates indicate that masks generated with different scoring seeds produce contrasting structures.  For example, the similarity matrix in Figure \ref{fig:winningtix} and the summary statistics in Table \ref{statistics} show all algorithm combinations yield \gls{ji} averages less than 0.29, a large difference between masks in all circumstances.  \gls{smc} values yielded higher scores (up to 0.84 in Table  \ref{statistics}), but this is expected at higher prune rates as most masks will have matching zeroes.  A telling distinction of uniqueness is the \gls{ji} at higher prune rates: at 90\% pruning, there is very low commonality between positive masks chosen for the subnetworks (0.05 and 0.09).

Results also indicate several similarities.  Figure \ref{galore1} compares each model against 14 others for each algorithm, with \gls{ji} scores falling within $1/100$ of a decimal place of each other.  We speculate this as being a result of the algorithm and the prune rate: since each layer is confined to a specific prune rate, its match rate compared to the same layer in another model will be constrained by the prune rate.   A second similarity can be seen in layerwise coefficients yielding similar patterns across models. The first layer exhibits the highest similarity in all cases, indicating the masking structure needed to learn dataset inputs is important.  Middle layers yielded diverse masks, showing that weights at these layers exhibit more interchangeable utility.   Finally, the last layer generally yielded higher similarity coefficients, indicating the importance of specific weights for classification.

Practically, this analysis provides several avenues for future research.  First, the existence of MPT's under hyperparameter control shows evidence that the total quantity of MPT's may be large.  Understanding this theoretical quantity may guide us in the search for better models.   And second, assessing the similarities of weights across MPT's can help us in understanding the desirable properties inherent to successful subnetworks.
\vspace{-2pt}

\section{Related Work}
\noindent \textbf{Traditional Network Pruning}
The effectiveness of sparse neural networks was first demonstrated by Lecun et. al \cite{lecun_optimal_1989}.  With the advent of deep learning, the size and efficiency of ML models quickly became a critical limitation.  Naturally, research aimed at decreasing size \cite{han_learning_2015, hinton_distilling_2015}, and limiting power and energy consumption \cite{yang_designing_2017}.  

\vspace{-2pt}
\noindent \textbf{Lottery Ticket Hypothesis}
The Lottery Ticket Hypothesis found that dense networks contained randomly-initialized subnetworks that, when trained on their own, achieved accuracy comparable to the original dense model. However, the approach required training a dense network in order to identify winning tickets.  Subsequent work identified strategies to prune \gls{dnns} \textit{without} a pretrained model using greedy forward selection \cite{ye_good_2020}, mask distances \cite{you_drawing_2022}, flow preservation techniques \cite{wang_picking_2019,tanaka_pruning_2020}, and  channel importance \cite{wang_pruning_2019}.  

\noindent \textbf{Randomized Neural Networks}
Important to work described in \cite{ramanujan_whats_2020,chijiwa_pruning_2021,malach_proving_2020},
 randomized neural networks \cite{gallicchio_deep_2020} have also been explored in shallow architectures.  Several applications explore randomization, including random vector functional links \cite{needell_random_2020,pao_learning_1994,pao_functional-link_1992},  random features for kernel approximations \cite{le_fastfood_2013,rahimi_random_2007,hamid_compact_nodate}, reservoir computing \cite{lukosevicius_reservoir_2009}, and stochastic configuration networks \cite{wang_stochastic_2017}.  

\noindent \textbf{Binary Neural Networks} \gls{bnns} studied in this paper fall into the class of quantized neural networks.  Like pruning, quantization is a natural approach for model compression. 
Common techniques for creating quantized networks include post-training quantization with retraining \cite{gysel_ristretto_2018, dettmers_8-bit_2016} and quantization-aware training \cite{gupta}.  In the Biprop algorithm proposed in \cite{diffenderfer_multi-prize_2021}, quantization-aware binary neural networks are trained with parameter $\alpha$, which enables floating-point weights to learn a scale parameter prior to binarization   \cite{martinez_training_2020}.  



\section{Discussion}

In this work, we propose a novel algorithm for finding highly accurate subnetworks within randomly initialized models.  Iterative Weight Recycling is successful on both \gls{dnns} (Edge-Popup) as well as  \gls{bnns} (Biprop).  Our results indicate that smaller networks are able to achieve higher accuracy than previously thought.  Practically, this allows us to create accurate and compressed models in limited compute environments.  

In addition, we show evidence of abundant \gls{mpts} by creating variegated subnetworks with nearly identical hyperparameters.  This provides several avenues for further investigation:  1) Deriving the theoretical limits on the total number of \gls{mpts} in a given architecture 2)  Exploring the properties of the unpruned weights to better understand weight optimization, and 3) Exploring weight pruning in different problem domains such as NLP.

\bibliographystyle{icml2022}
\bibliography{weight_recycling}
\newpage
\newpage
\appendix
\onecolumn


\section{Analysis: Random Weights}
In Figure \ref{norms}, we analyze the norms of the weights chosen by various subnetwork identification algorithms.  We train each subnetwork identification algorithm on Kaiming Normal randomly initialized weights with a 50\% prune rate.  

In Figure \ref{norms}, we measure the Frobenius norms of each layer for the unpruned masks (denoted with a "+"), as well as the norms of weights for the pruned masks (denoted with a "-").  Unpruned masks ("+") are the identified subnetworks chosen by each algorithm.   Results show that for each of the four algorithms, the weights at positive masks contain higher norms than the weights of disposed zero masks.  Notably, the two highest identified norms, IteRand and Weight Recycling, perform the strongest.  Weight Recycling negative masks overlap with its positive masks almost identically, and Edge-Popup and Biprop negative masks overlap almost identically with each other.  

This shows evidence that high norm weights are beneficial to a successful subnetwork in each algorithm, with Iterative Weight Recycling emphasizing the reuse of these high norm weights across its pruned and unpruned weight masks.  

\begin{figure}[!h]
\begin{center}

   \includegraphics[scale=0.75]{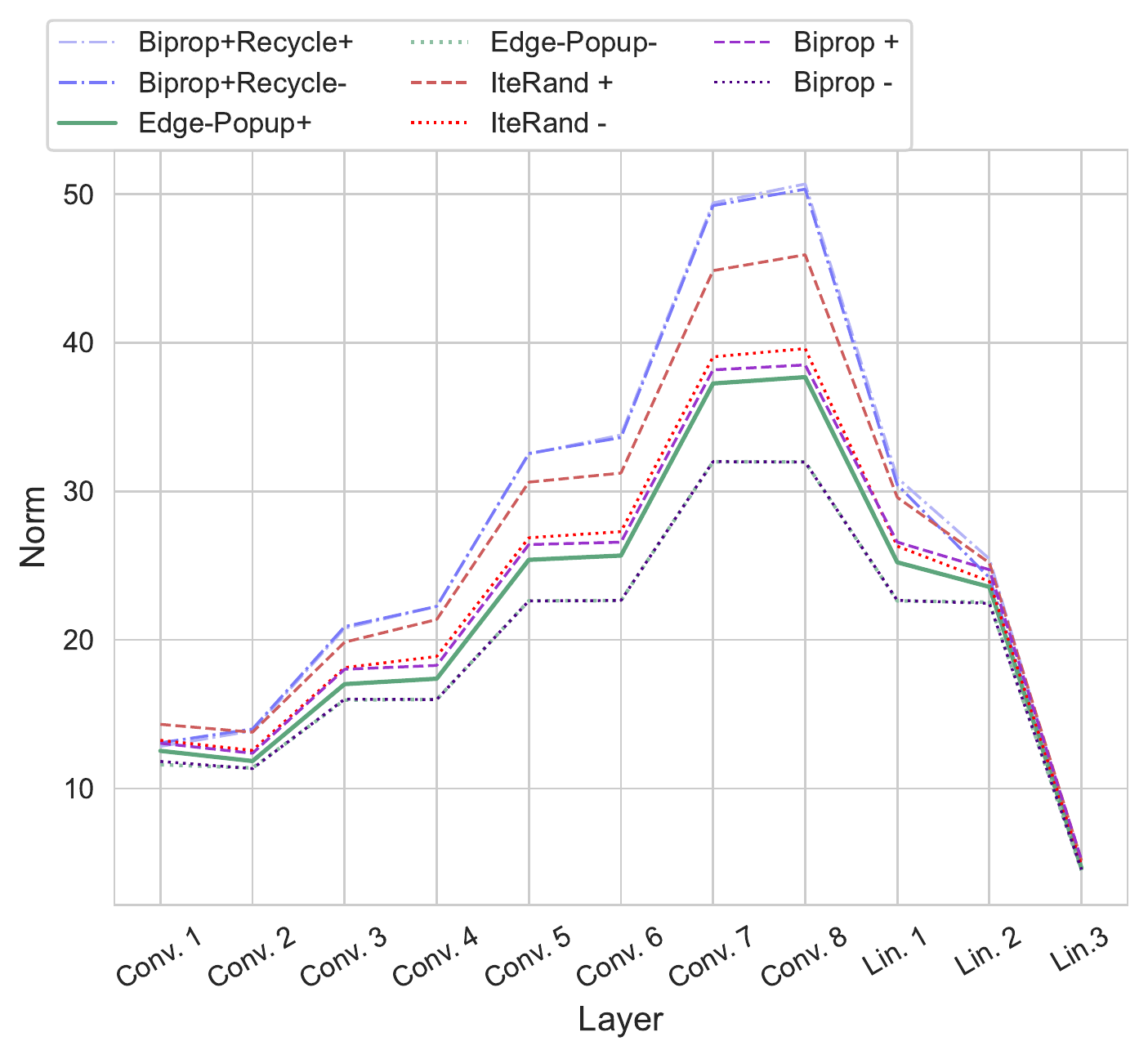}
    \caption{  Frobenius norms across algorithms and model layers for pruned and unpruned weights. "+" denotes the learned high-accuracy subnetwork, "-" denotes the pruned weights.  Each algorithm keeps weights with higher norms, discarding lower norm weights.  Weight Recycling reuses high norm weights to identify the best subnetwork.     }\label{norms}
    \centering
\end{center}
\end{figure}

\section{CIFAR-10 Hyperparameters}
For CIFAR-10 experiments, we test each algorithm using Conv-2 to Conv-8 architectures as well as ResNet18.  Additionally, we test each algorithm with layer width parameter $p$ for Conv-2 to Conv-8 algorithms.  These architectures are built using the same codebase as Edge-Popup and IteRand.  We refer to the Edge-Popup GitHub for model definitions: \newline
\href{https://github.com/allenai/hidden-networks/tree/master/models}{https://github.com/allenai/hidden-networks/tree/master/models}

\noindent \textbf{Baseline models}
For our baseline models, we train dense models with learned weights.  We use the SGD optimizer initialized with kaiming normal weights, with a learning rate of 0.01 for 100 epochs with batch size 128.  We additionally use a weight decay of 1e-4 and momentum 0.9.  Cosine decay learning rate policy is used for all models. 

\noindent \textbf{Subnetwork algorithms}
We train Edge-Popup, IteRand, Biprop, and Weight Recycling with similar hyperparameters.  We try to train baseline algorithms with the same hyperparameters as used in the original papers (ass denoted below).  

All models are are trained with hyperparameters as follows: \vspace{-1em}
\begin{enumerate}
  \setlength{\itemsep}{1pt}
  \setlength{\parskip}{0pt}
  \setlength{\parsep}{0pt}
  \item SGD optimizer, learning rate 0.1
  \item 250 Epochs
  \item Batch Size=128
  \item Weight Decay=1e-4 for Conv2-Conv8 models, 5e-4 for ResNet
  \item Cosine decay learning rate policy
  \item Non-Affine BatchNorm
  \item Score parameter $\mathrm{S}$ initialized with Kaiming Uniform
\end{enumerate}
\vspace{-1em}

\begin{table}
\begin{center}
\renewcommand{\arraystretch}{1.15}
\begin{tabular}{|c| c| } 
 \hline
 Algorithm & Weight Init.  \\ [0.5ex] 
 \hline
 Edge-Popup& Signed Constant \\ 
 Edge-Popup+IteRand &  Signed Constant \\

 Biprop & Kaiming Normal\\
 
 Biprop+IteRand &  Kaiming Normal \\
 Edge-Popup+Weight Recycle &Signed Constant \\
 Biprop+Weight Recycle &  Kaiming Normal \\

 \hline
\end{tabular}
\caption{Weight initializations used for each experiment.  }
\end{center}
\end{table}

\noindent \textbf{IteRand with Biprop}
We try several configurations for Biprop + IteRand, applying the same code as the original authors from each paper.  Configurations  
Signed Constant and Unsigned Constant initialization, modified learning rate, modified rerandomization frequency and rate (we chose parameters identical to Weight Recycling, 10 and 0.2, respectively), modified score seed, modified weight seed.  In the result tables below, we use the same hyperparameters as the original paper, using Kaiming Normal initialization for Biprop+IteRand experiments.  

\noindent \textbf{Model Sizes}

\begin{center}
\renewcommand{\arraystretch}{1.15}
\begin{tabular}{c c c c c c c} 
 Prune Rate & Width Factor & Conv2 & Conv4 & Conv6 & Conv8 & ResNet18 \\ [0.5ex] 
 \hline\hline
 0 & 1 & 4,300,992&2,425,024&2,261,184&5,275,840&11,678,912 \\ 
 \hline 
  0 & 0.1 &39,761&22,505&21,630&51,614&- \\ 
 \hline
   0 & 0.25 & 269,616&152,368&142,128&330,032&- \\ 
 \hline
   0 & 0.5 &1,076,320&607,328&566,368&1,319,392&-  \\ 
 \hline
 0.2 & 1 & 3,440,794 & 1,940,019 & 1,808,947 & 4,220,672 & 9,343,129.60\\
  \hline
 0.4 & 1 & 2,580,595 & 1,455,014 & 1,356,710 & 3,165,504 & 7,007,347.20\\
  \hline
 0.5 & 1 & 2,150,496 & 1,212,512 & 1,130,592 & 2,637,920 & 5,839,456.00\\
  \hline
 0.6 & 1 & 1,720,397 & 970,010 & 904,474 & 2,110,336 & 4,671,564.80\\
  \hline
   0.8 & 1 & 860,198 & 485,005 & 452,237 & 1,055,168 & 2,335,782.40\\
  \hline
 0.9 & 1 & 430,099 & 242,502 & 226,118 & 527,584 & 1,167,891.20\\
  \hline
 0.95 & 1 & 215,050 & 121,251 & 113,059 & 263,792 & 583,945.60\\
  \hline
 0.98 & 1 & 86,020 & 48,500 & 45,224 & 105,517 & 233,578.24\\
  \hline
 0.99 & 1 & 43,010 & 24,250 & 22,612 & 52,758 & 116,789.12\\
  \hline
 0.5 & 0.1 & 19,881 &11,253  & 10,815 & 25,807 & -\\
  \hline
 0.5 & 0.25 & 134,808 & 76,184 & 71,064 & 165,016 & -\\
  \hline
 0.5 & 0.5 & 538,160 & 303,664 & 283,184 & 659,696 & -\\
 \hline

 \hline
\end{tabular}
\end{center}

\section{CIFAR-10 Results}
In this section we detail the results across CIFAR10 models.  Accuracies for Weight Recycling experiments are averaged across three runs.

Dense model accuracy's:
\begin{table}[!h]
\begin{center}
\begin{tabular}{c c c c c c c} 
 Prune Rate & Width Factor & Conv2 & Conv4 & Conv6 & Conv8 & ResNet18 \\ [0.5ex] 
 \hline\hline
 0 & 1 &79.9 & 87&89 &89.41 & 93.03 \\ 
 \hline 
  0 & 0.1 & 58.5&67.26 & 72.48& 75.06& -\\ 
 \hline
   0 & 0.25 &72.49 & 81.13& 83.65& 84.75& -\\ 
 \hline
   0 & 0.5 &77.55 &84.96 &87.54 &87.18 &-  \\ 
 \hline
 \hline
\end{tabular}
\end{center}
\end{table}

\begin{table}[!h]
\begin{center}
\begin{tabular}{c c c c c c c } 
 Prune Rate & Width Factor &Algorithm& Conv2 & Conv4 & Conv6 & Conv8  \\ [0.5ex] 
 \hline\hline
\multirow{ 3}{*}{0.2} &\multirow{ 3}{*}{1} & Biprop & 64.1&74.71&78.2&76.59 \\
&& IteRand & 56&57.9&59.6&49.84 \\
&& Weight Recycle& \textbf{80.28}&\textbf{88.05} &\textbf{90.2}&\textbf{90.35} \\
\hline

\multirow{ 3}{*}{0.4} &\multirow{ 3}{*}{1} & Biprop &78.3 &85.9&88.9& 89.13\\
&& IteRand & 65& 73.3&75& 75.14\\
&& Weight Recycle&\textbf{81.2} & \textbf{88.8}&\textbf{90.8}& \textbf{90.93}\\
\hline

\multirow{ 3}{*}{0.5} &\multirow{ 3}{*}{1} & Biprop & 79.56&87.42&89.52&90.35 \\
&& IteRand &65.3 &74&77& 77.61\\
&& Weight Recycle&\textbf{81.36 }&\textbf{88.91}&\textbf{90.9}&\textbf{ 91}\\
\hline

\multirow{ 3}{*}{0.6} &\multirow{ 3}{*}{1} & Biprop & 79.98&87.2&90.25& 90.54\\
&& IteRand & 65.72& 74.13&76.2& 78.9\\
&& Weight Recycle&\textbf{81.36 }&\textbf{88.64}&\textbf{90.6}&\textbf{90.85} \\
\hline

\multirow{ 3}{*}{0.8} &\multirow{ 3}{*}{1} & Biprop &77.9 &85.34&87.5&89.11 \\
&& IteRand &58.9 & 67.1&71.2& 76.72\\
&& Weight Recycle&\textbf{ 80.23}&\textbf{ 87.75}&\textbf{90}&\textbf{ 90.55}\\
\hline

\multirow{ 3}{*}{0.9} &\multirow{ 3}{*}{1} & Biprop & 70.1&79&82.38& 84.84\\
&& IteRand &50 & 54&61&52 \\
&& Weight Recycle&\textbf{76.9} &\textbf{84.3} &\textbf{85.5}&\textbf{ 86.4}\\
\hline

\multirow{ 3}{*}{0.95} &\multirow{ 3}{*}{1} & Biprop & 56.6&64.83&66.12&- \\
&& IteRand & 38.78& 45.91&49.2&- \\
&& Weight Recycle&\textbf{65.11} &\textbf{73.5 }&\textbf{72.11}& -\\
\hline

\multirow{ 3}{*}{0.5} &\multirow{ 3}{*}{0.1} & Biprop &47.2 &50.4 &56.9 & 61.2 \\
&& IteRand & 41 &42.8 &43 & 44.3  \\
&& Weight Recycle& \textbf{52.55}& \textbf{59.64}&\textbf{65}& \textbf{69} \\
\hline

\multirow{ 3}{*}{0.5} &\multirow{ 3}{*}{0.25} & Biprop &64.55 & 73.23& 77.84&  80.4\\
&& IteRand & 53 & 57.5&60.4 & 62.5  \\
&& Weight Recycle& \textbf{68.8}& \textbf{78.51}& \textbf{82}& \textbf{84.5} \\
\hline

\multirow{ 3}{*}{0.5} &\multirow{ 3}{*}{0.5} & Biprop & 74.69&82.6 & 86.23&  87.33\\
&& IteRand & 60.5 &67.3 &70.4 &70.3   \\
&& Weight Recycle&\textbf{76.7} & \textbf{86.5}&\textbf{88.43} &\textbf{89.67}  \\
\hline

\hline
 \hline
\end{tabular}

\end{center}
\caption{\textbf{Binary neural networks}
Biprop, Biprop+IteRand,Biprop+Iterative Weight Recycling results. Weight recycle accuracy averaged across three runs. Best result bolded for each experiment.  Notably, Weight Recycling performs best across all algorithms, and is robust to higher prune rates.  }
\end{table}

\begin{table}[!h]
\begin{center}
\begin{tabular}{c c c c c c c } 
 Prune Rate & Width Factor &Algorithm& Conv2 & Conv4 & Conv6 & Conv8  \\ [0.5ex] 
 \hline\hline
\multirow{ 3}{*}{0.2} &\multirow{ 3}{*}{1} & Edge-Popup & 70.1&77.6&83&84.39 \\
&& IteRand & 80.7&88.4 &90.58&90.99 \\
&& Weight Recycle & \textbf{81.05}&\textbf{ 88.95}&\textbf{91.1}&\textbf{91.15} \\
\hline

\multirow{ 3}{*}{0.4} &\multirow{ 3}{*}{1} & Edge-Popup & 78.6 &86.74&89.33&90.15\\
&& IteRand & 81.3 &89.2&91.1&\textbf{91.55}\\
&& Weight Recycle& \textbf{81.8}& 89.2&\textbf{91.2}&91.5 \\
\hline

\multirow{ 3}{*}{0.5} &\multirow{ 3}{*}{1} & Edge-Popup & 79.69 &86.67&89.53&90.28\\
&& IteRand & \textbf{81.83} &88.37&90.74&91.11\\
&& Weight Recycle& 81.68&\textbf{88.5} &\textbf{90.9}& \textbf{91.36}\\
\hline
\multirow{ 3}{*}{0.6} &\multirow{ 3}{*}{1} & Edge-Popup & 79.1 &86.43&89.1&90.5\\
&& IteRand & \textbf{81.75} &87.66&90&\textbf{91.16}\\
&& Weight Recycle&81.12 & \textbf{88.15}&\textbf{90.4}&90.95 \\
\hline
\multirow{ 3}{*}{0.8} &\multirow{ 3}{*}{1} & Edge-Popup & 75.15 &83.28&85.4&86.9\\
&& IteRand & 78.04 &85.34&86.94&88.42\\
&& Weight Recycle&\textbf{79.06} &\textbf{ 85.6}&\textbf{87.13}&\textbf{88.87} \\
\hline
\multirow{ 3}{*}{0.9} &\multirow{ 3}{*}{1} & Edge-Popup & 64.05 &74.42&79.57&83.6\\
&& IteRand & 68.87 &78.88&81.22&84.73\\
&& Weight Recycle&\textbf{70.22} &\textbf{79.78} &\textbf{83.2}&\textbf{86.93} \\
\hline
\multirow{ 3}{*}{0.95} &\multirow{ 3}{*}{1} & Edge-Popup & 49.4 &57.61&60.63&76.7\\
&& IteRand & 53.5 &64.72&\textbf{68.8}&\textbf{81.66}\\
&& Weight Recycle& \textbf{55.2}& \textbf{65.22}&68.02&81.51 \\
\hline
\multirow{ 3}{*}{0.5} &\multirow{ 3}{*}{0.1} & Edge-Popup & 48.1 &50.73&57.3&63.35\\
&& IteRand & \textbf{55.2} &59.3&\textbf{64.5}&\textbf{71.56}\\
&& Weight Recycle& 53.4&\textbf{59.82} &64.1&70 \\
\hline

\multirow{ 3}{*}{0.5} &\multirow{ 3}{*}{0.25} & Edge-Popup & 72.49&73.34&78.48&81.5\\
&& IteRand &  \textbf{70.5}&78.91&83&85.5\\
&& Weight Recycle& 70.33& 78.92&83& 85.5\\
\hline

\multirow{ 3}{*}{0.5} &\multirow{ 3}{*}{0.5} & Edge-Popup & 77.55&82.08&86.24&87.92\\
&& IteRand &  \textbf{78}&85.45&88.18&87.99\\
&& Weight Recycle& 77& \textbf{85.63}&\textbf{88.57}&\textbf{89.73} \\
\hline

\hline
 \hline
\end{tabular}

\end{center}
\caption{\textbf{Randomly Initialized Neural Networks}
Edge-Popup, Edge-Popup+IteRand, Edge-Popup+Iterative Weight Recycling results. Weight recycle accuracy averaged across three runs.  Best result bolded for each experiment. Notably, at prune rates above 80\%, weight recycling outperforms all algorithms.  }
\end{table}

\begin{figure}[!h]
   \includegraphics[width=1.0\textwidth,height=4.5cm]{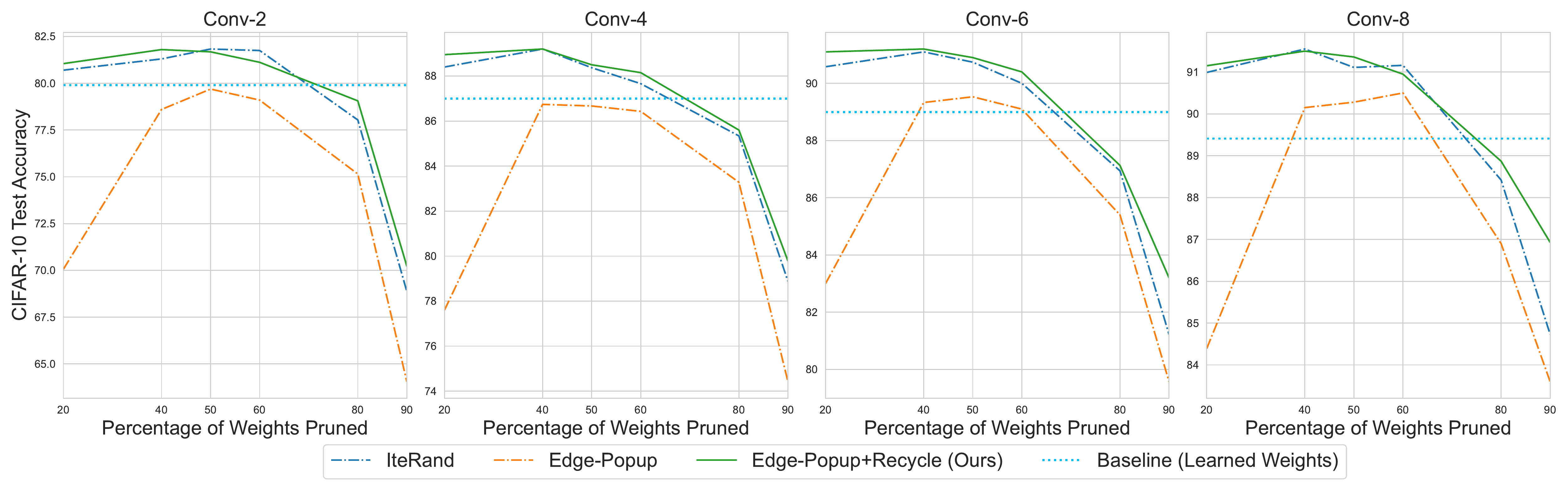}
    \caption{Comparison of Edge-Popup, IteRand, and Weight Recycling using continuous valued weights with signed constant initialization. In this figure, we vary the model depth. }\label{depth_bin}
    \centering
\end{figure}
\vspace{1em}

\begin{figure}[!h]
   \includegraphics[width=1.0\textwidth,height=4.5cm]{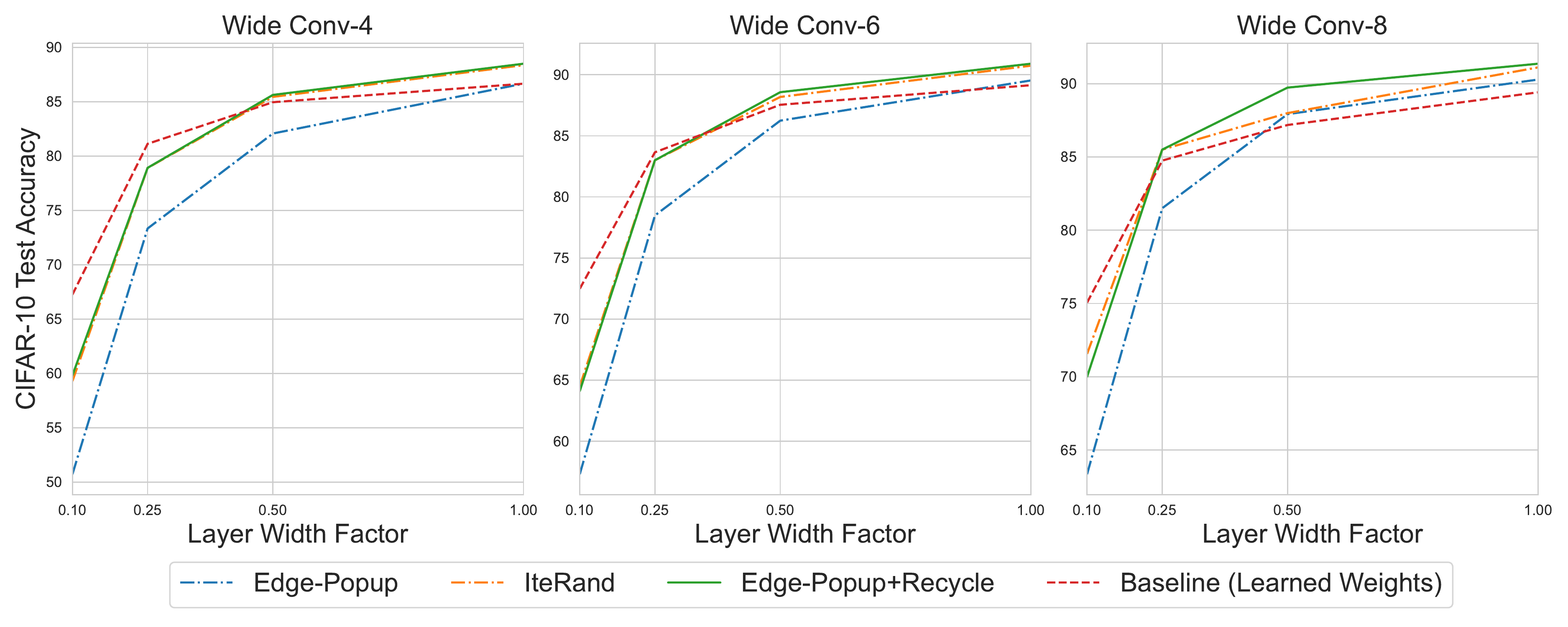}
    \caption{Comparison of Edge-Popup, IteRand, and Weight Recycling using continuous valued weights with signed constant initialization. In this figure, we vary the model width. }\label{depth_bin}
    \centering
\end{figure}

\section{MPT Analysis}
In Section "Winning Tickets Galore", we show that an abundance of \gls{mpts} (winning tickets) exist in randomly
initialized neural networks.  We analyze \gls{mpts} in both randomly initialized neural networks using Edge-Popup as well as binary randomly initialized networks using Biprop.  We train 15 Edge-Popup models and 15 Biprop models with a Conv-6 model at 50\% prune rate, and 5 models of each algorithm at both a 75\% and 90\% prune rate (using Conv-6).  

We restrict the hyperparameters of each run to measure the differences of the generated subnetworks.  The only hyperparameter that is changed is the score parameter seed, which is set to values 0 to 14 for 50\% prune rate, and 0 to 4 for 75\% and 90\% prune rates.  The rest of the hyperparameters are defined as follows: 

\begin{enumerate}
  \setlength{\itemsep}{1pt}
  \setlength{\parskip}{0pt}
  \setlength{\parsep}{0pt}
  \item SGD optimizer, learning rate 0.1
  \item Weight decay 1e-4, momentum 0.9
  \item 250 Epochs, Batch size 128
  \item Batch Size=128
  \item Weight initialization: Signed constant for Edge-Popup, Kaiming Normal for Biprop
  \item All random seeds set to 0: python random library, torch manual seed, torch cuda manual seed, torch cuda manual seed all
  \item torch.backends.cudnn.deterministic = True
  \item torch dataloader: worker\_init\_fn=np.random.seed(0)
  \item Weight Seed=0

\end{enumerate}

We measure the \gls{smc} and \gls{ji} of each model combination.  The figures below summarize the \gls{smc} and \gls{ji} for combinations at 50\% and 75\% prune rate.  

\begin{figure}[h]
\begin{center}

   \includegraphics[scale=0.25]{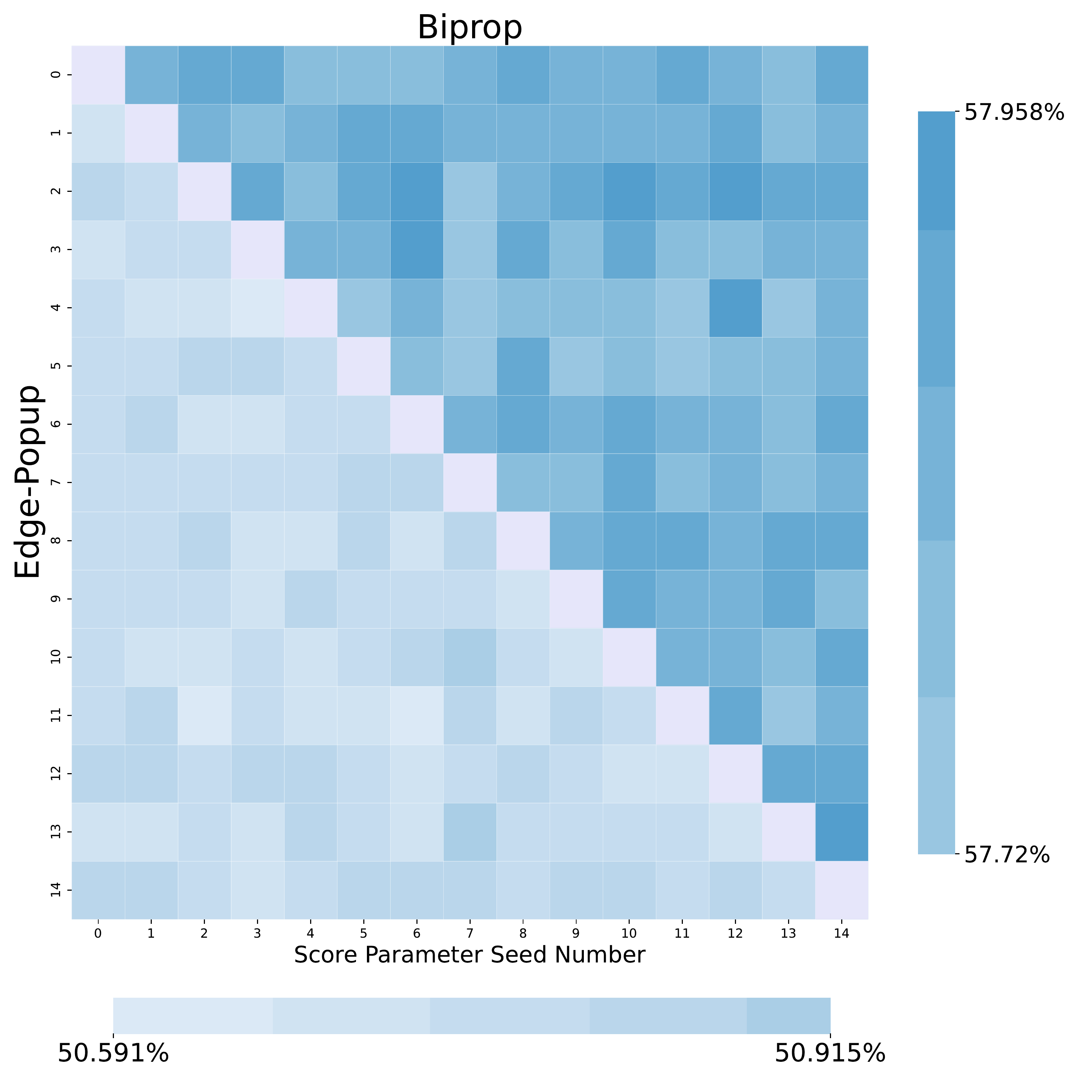}
    \caption{ Heatmap of Simple Matching Coefficients of Binary Masks across models.   }\label{depth_bin}
    \centering
\end{center}
\end{figure}

\begin{figure}
\centering
\begin{minipage}{.5\textwidth}
  \centering
  \includegraphics[width=.4\linewidth]{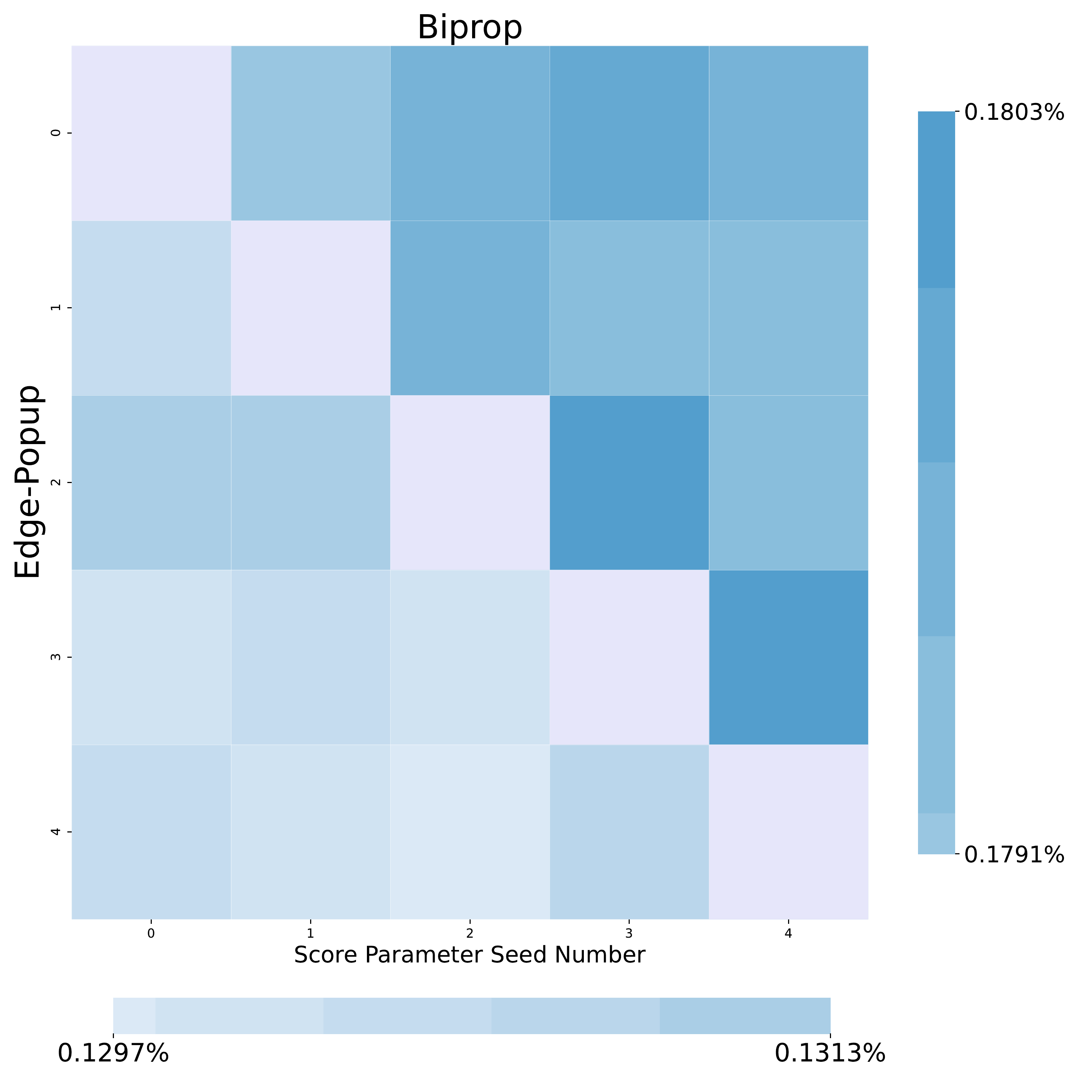}
  \caption{Jaccard Index at 75\% prune rate.}
\end{minipage}%
\begin{minipage}{.5\textwidth}
  \centering
  \includegraphics[width=.4\linewidth]{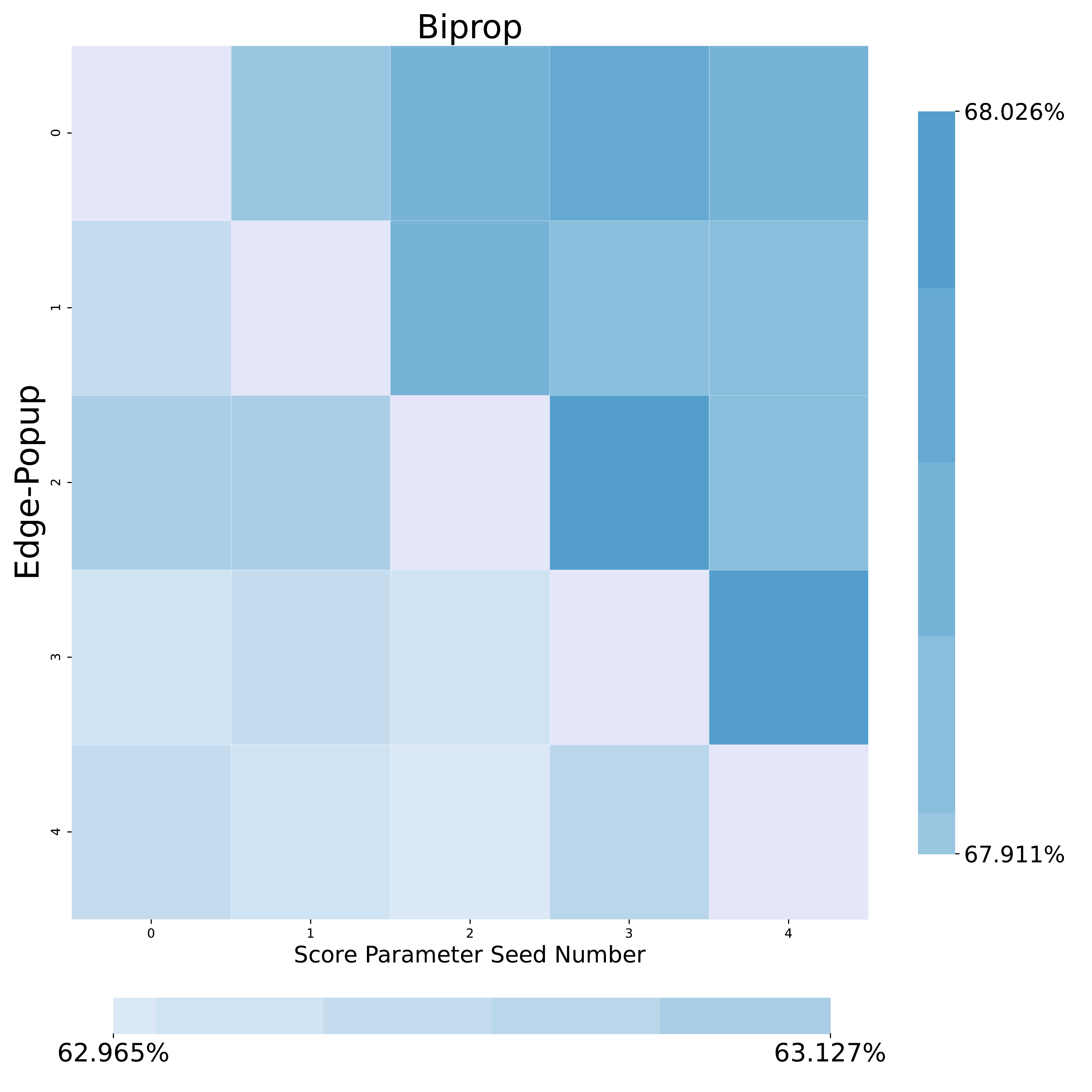}
  \caption{SMC at 75\% prune rate.}
\end{minipage}
\end{figure}

\end{document}